# An Overview of Artificial Intelligence-based Soft Upper Limb Exoskeleton for Rehabilitation: A Descriptive Review

Sanjukta Halder[1], Dr. Amit Kumar[2]

[1,] Faculty, ASDC Solutions, Kolkata

[2] Professor, ME Dept, NIT Patna

**Abstract:** The upper limb robotic exoskeleton is an electromechanical device which use to recover a patient's motor dysfunction in the rehabilitation field. It can provide repetitive, comprehensive, focused, positive, and precise training to regain the joints and muscles' capability. It has been shown that existing robotic exoskeletons are generally used rigid motors and mechanical structures. Soft robotic devices can be a correct substitute for rigid ones. Soft exosuits are flexible, portable, comfortable, user-friendly, low-cost, and travel-friendly. Somehow, they need expertise or therapist to assist those devices. Also, they cannot be adaptable to different patients with non-identical physical parameters and various rehabilitation needs. For that reason, nowadays we need intelligent exoskeletons during rehabilitation which have to learn from patient's previous data and act according to it with patient's intention. The exoskeletons have to the power of decision-making to avoid the presence of expertise. There also has a big gap between theoretical and practical applications for using those exoskeletons. Most of the intelligent exoskeletons are prototype in manner. To solve this problem, the robotic exoskeleton should be made both criteria as ergonomic and portable. In this growing field, the present trend is to make the exoskeleton intelligent and make it more reliable to use in clinical practice.

**Introduction:** Now-a-days stroke is a very common disorder of blood circulation in human arteries; It obstructs the flow of the vessel. Temporary or permanent disability may occur in adults because of these conditions [1] [2]. Currently, 60% of stroke patients are belongs to India in the world as per India express. Approximately 4000 stroke cases happen daily in India and not more than 2-3% are treated, among them 50-80% are in the acute phase and 40-50% are in the chronic phase. After a stroke, rehabilitation must be needed, it can prevent the next stroke which may occur within five years after the first stroke. Traditional treatments such as physiotherapy use simple equipment or monitoring to help patients return to normal life with some uncertainty, even can be limited by wheelchairs [3][4]. Expertise gives an indication that robot-assisted rehabilitation can improve the motor function of a patient's limbs with appropriate treatment [5].

Exoskeletons start a new period of the modern neuromuscular rehabilitation process and research in assistive technology [6]. Traditional exoskeletons are rigid and heavy as per previous publications, which may be eliminated by soft exoskeletons in future, which has been developed in the last decade and its development is gradually increasing day by day with further research. Soft exoskeletons are able to decrease the weight of the devices due to its light nature. The rehabilitation intelligent robotic device can record patient data and estimate exercise capability due to high-level control techniques that can be implemented in a range of methods [7]. The rehabilitation robots communicate with patients directly, thus the robots acquire patient data easily and protect it with effective training. In Addition, the rehabilitation process can be more fruitful with virtual reality or other techniques such as gaming technology [8]. Artificial intelligence (AI) presents some algorithms which acquired knowledge from occurrences and forecast exotic conditions, which increases robot-assisted rehabilitation intelligence [9]. Robots can analyze native movements and compute human desires combined with AI. Intelligent systems can acquire knowledge from past happening processes and robotic models can adapt gracefully to dynamic and unknown environments by following these processes [10][11].

Although existing review articles discussed robotic sensing technology, degree of freedom, computational techniques and human-robot interactive control methods, intention-detection techniques, quantitive assessment etc. The benefits and drawbacks of every method are also presented by doing these studies. So, the representative's investigation left some concerns about AI-based robot-assisted rehabilitation.

The aim of this research is to summarize AI-based soft wearable robotic devices and assistance with a focus on upper limb rehabilitation with intelligence. Over the last two decades, robotics technology has been used in rehabilitation and many devices have been proposed depending on this topic. There are many devices that do not have clinical acceptance due to their poor therapeutic utility, many of them were not transportable, airy and convenient. Low-to-middle-income countries cannot adopt exosuits due to high cost whereas stroke happens here more than in high-income countries [12]. In this way, various strategies with AI have to be applied to reduce the cost of production.

**Methods:**

A smart soft exoskeleton is a recent trend in this field, with only 3% to 4% of AI-based exoskeletons being in soft exoskeletons. As soft exoskeletons have gained attention over the past few decades, no such device has yet been presented on the market. Here we present a descriptive review of intelligent soft exoskeletons.

Various papers were searched through search engines for this review. Among these sources, including IEEEXplore, PubMed, Scopus, Web of Science, etc., only very few papers met the criteria. The screening and qualification process is accomplished by the PRISMA flow diagram. In order to obtain a suitable literature review the investigated papers should have the following characteristics listed below:

Eligibility Criteria

- Papers should be written in English language only.
- The exoskeleton must be positioned with a minimum of one degree of freedom from upper extremities such as shoulders, wrists, elbows, and hands.
- The device should be portable, wearable, and lightweight.
- Exosuit must have at least one artificial intelligence algorithm used.
- Desirable exoskeletons are used for rehabilitation, support, and augmentation applications.

- The executed intention-detection techniques for upper limb exosuit.

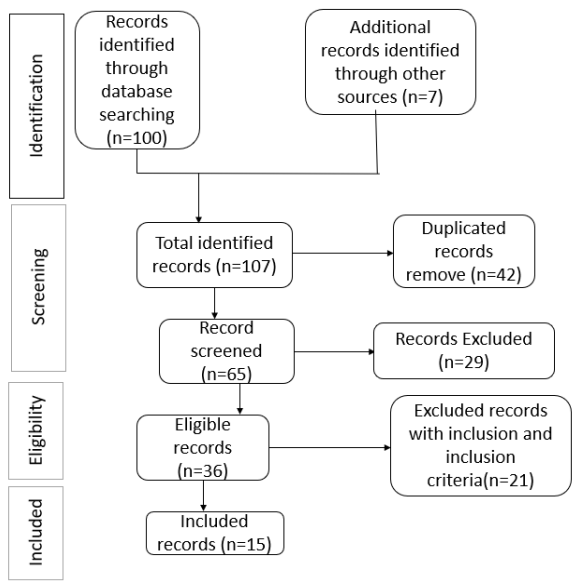

Figure1: PRISMA flow diagram for the processing of the survey

**Results:**

About 100 papers were identified by keyword through a digital database, analyzing a small number of articles that followed the criteria. Some articles are not written in English. Some exoskeletons use AI techniques but are not soft in structure. Most soft exoskeletons are using model-based control methods. So, keeping all these in mind, the selection process is done through the flowchart shown in Figure 1. This survey has been created following all the sins.

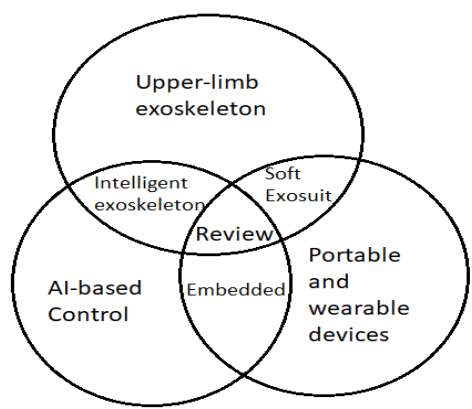

**Figure 2: Graphic summary of the descriptive review**

In general, a graphical framework is shown in Figure 2. Here it can be seen that the intelligent soft exoskeleton is actually a combination of different fields. Out of 100 papers, only 15 papers on intelligent soft exosuits were found after qualification and analysis. A descriptive review is presented on their synthesis.

I. Recap of Upper-Limb Soft Exoskeletons:

An exoskeleton of the upper limb describes as the covering device of the human arm. Development of soft exoskeletons has increased over the last decade, particularly in lower limb applications. Soft exoskeletons may replace many or all hard exoskeletons in the future due to light, thin and flexible materials are the substitute for hard, heavy, and rigid materials for soft exoskeletons [13]. The flexible soft exoskeleton is also known as an Exosuits. Some components must be rigid, such as the battery and controller, which are often placed in a bag or separately somewhere to reduce weight [14] and make the suit ergonomic. That types of exosuits are great because of their features as they are easy to transport and install. Patients can use these devices alone and carry them around. Therefore, they are user-friendly and can be used in daily life for improvement easily. Wearable soft exoskeleton robots have gradually become popular in clinics because they can support multiple joints and provide flexible and safe rehabilitation training therapy. Most of the rehabilitation exoskeletons are in the prototype stage still now [15]. Upper limb exoskeletons work parallel to the human arm and are attached to the human upper limb at various joints. By this indication, it is clear that robots have to adapt to different arm lengths.

Some examples of soft exoskeletons are ; Chiaradia et al [16] proposed a refined design of the exosuit of which a smooth control architecture represented motion-intention detection with gravity compensation for aiding elbow. Oguntosinet et al [17] representation, he revealed a prototype of an inexpensive, airy, and wearable soft robotic-aided device that could assist a patient with elbow motion therapies after a stroke occurs. Suresh Gobee et al [18] proposed a portable soft robot assistive device for finger rehabilitation. It utilized electromyography (EMG) signals to catch patients' muscular signals, amplified those signals, and simulated solenoid valves for controlling pneumatic actuators attached to the wearable gloves. Yet Zhang et al [19] demonstrated a conformal mechanism of a multi-material pneumatic soft finger that was optimized to obtain its maximum bending deflection and also modified for artificial hand gripper, rehabilitation and other practical applications. Chingyi Nam, et al. [20] elaborated lightweight, ergonomic and low power-demanding soft wearable robot for the blooming system can aid the elbow, wrist, and fingers to fulfil the following arm reaching in a logical order and withdrawing through EMG signals.

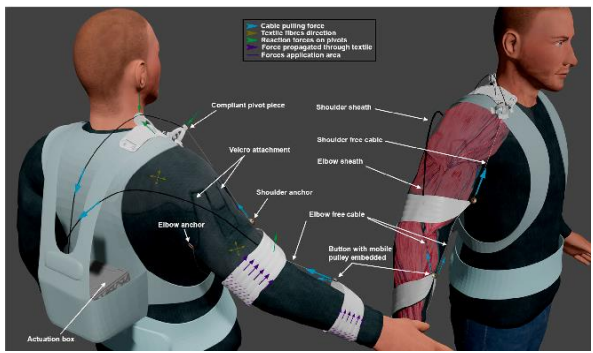

**Figure 3: Example of soft exosuit © [14]**

Mainly most soft exosuits are made of fabric or textiles or elastomers that bundle around the user's limbs and connect activities to the limbs [21].

II. Actuated Joints with DOFs:

Soft exosuits can be used for the treatment of the upper limb, including the actuated joints of the shoulder [14,22], wrist [23,25], elbow [16,17], and hand and finger [18,19] aid. The DOF determines the possible movement of a limb over one or different joints. The design of an exoskeleton is complex with higher DOFs and needed much improvement in control techniques to do the specific task. The design is simple to use with one DOF [26] [27] [28] [29] [30], and the driving mechanism allows accurate joint mobility but increases the complexity of the controller. Approximately 58% of research considered single DOF exosuits mostly based on the elbow joint [26] which assists in flexion and extension movements. The internal and external design of two DOFs robots [24] [18] are more complex than single. Almost 29% of reviewed papers are based on two DOF as shown in figure 4. All of those shoulder-elbow models are very common [17]. The robotic structure decides to improve the efficacy of rehabilitation guidance according to the patient's condition. This study presented 10% of reviews are on the basis of three DOFs models [19] [31] [32]. Very small researches are based on multi DOFs, but we need it more.

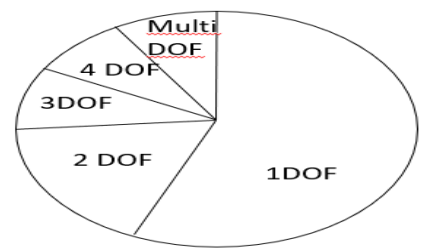

**Figure 4: General distribution of physical properties of robotic exosuits based on degrees of freedom.**

The shoulder is a vital joint for the upper limb, it is a kind of ball-socket joint. The disabilities of the shoulder can affect the whole upper limb badly. The shoulder joint has three degrees of freedom (DOFs) such as flexion-extension in the sagittal plane, abduction-adduction in the coronal plane, and medio-lateral rotation in the transverse plane. Elbow joint is a kind of hinge joint between the arm and forearm of a human body. It possesses two degrees of freedom as flexion-extension and supination- pronation to the forearm. The wrist joint is a complicated joint that bridges the gap between the hand and the forearm. It has three degrees of freedom - flexion-extension, radioulnar-deviation, and rotation. Finger joint dysfunction is a very crucial problem due to its most complicated anatomical nature. The fingers have 21 degrees of freedom (DOFs), including 5DOFs for the thumb and the remaining four fingers with 4 DOFs each for the index, middle, ring, and little fingers.

The characters of the exoskeletons depend upon the type of degrees of freedom (DOFs). In [22], Varghese and others represent a soft exosuit for the shoulder joint with two degrees of freedom for the motion in the sagittal and coronal plane and one degree for the elbow joint with the help of a bio-inspired kinematic sensing strategy. whereas SAMPER-ESCUDERO et al [14] designed a wearable exosuit made of elastic fabric that supports shoulder and elbow stretch. This soft prototype can support the elbow fully and the shoulder partially up to 100 degrees. This prototype suit contains two DOFs, one for shoulder joint flexion and one for elbow flexion. In a paper, Benjamin et al [23] spoke for a 3D-printed soft robotic wrist device for rehabilitation with the mobility of two DOF. By using this at least 70% recovery is possible with sufficient torque and bending.

Multi DOFs for upper limb exosuits lag behind due to complex control mechanisms. Multiple degrees of freedom are required for daily activities such as reading, drinking, eating, etc. through the soft exoskeleton after a stroke.

### III. Actuation and Transmission technique:

First step of design a soft exoskeleton is joints for movements and second step is choice of actuators. A variety of actuators can be used for making exosuits but according to research status, mostly used actuation systems are cable-driven modules, pneumatic actuators, shape memory alloy (SMA)-based actuators, and hybrid actuators. The cable-driven actuator uses cables and sheaths to transfer the power [33], where sheaths are usually installed on the surface of the body and follow the correspondence path from the prime mover to the anchor joint. In a pneumatic actuator, the volume or shape is changed according to the adjusting internal pressure and executes such actions as elongation, bending, or shortening [34]. But most of the pneumatic actuator-driven exosuits need external connections for heavy air sources, thatswhy they are not unportable. Shape memory alloy-based actuator[35] is very popular as they are lightweight, produce noiseless actuation. Copaci et.al. [36] represent an exoskeleton using a nonconventional actuation system on Bowden cable, Teflon sheath, and SMA wire for elbow flexion. Passive actuators mainly consist of spring and elastic fabric, where quasi-passive actuators are using spring, elastic fabric and quasi-passive elements such as variable dampers or electromagnetic clutch [37]. Hybrid actuators are made by combining pneumatic actuators and wire-driven transmissions [38]. Another type of popular hybrid actuation system is pneumatic actuators and neuromuscular electrical stimulation (NMES), which offer great potential for devices in rehabilitation.

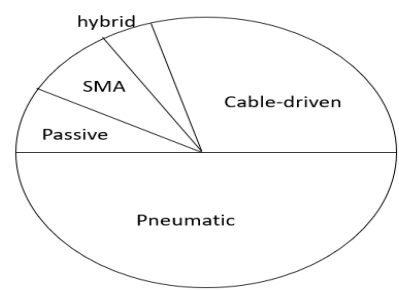

**Figure 5: general distribution diagram for actuation system**

The general distribution of actuation system is summarized in figure 5, where we can see that the published exosuits are used 32% cable-driven mechanisms, 50% publication utilized pneumatic actuation. SMA based actuators are used on in 7% and Passive actuation system in 7% publications. Among all past soft exoskeleton, only 4% are based on hybrid actuators.

IV. Sensing and Intention Detection:

During the rehabilitation process, the patient's needs for exercise greatly affect in recovery. Exercise's activeness increases the strength of central nerve system, this is why intention detection gives very important in human-robot interactive processing. Physiological and physical signals are goes through the artificial algorithms, specifically machine learning algorithms and get movement intention as shown in figure 6.

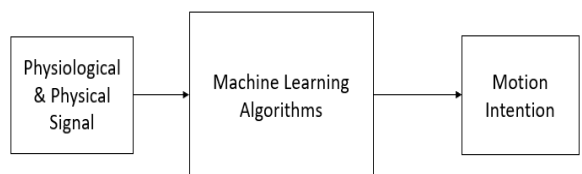

**Figure 6: schematic diagram of intention recognition**

Different types of physiological signals such as EMG (electromyogram), EEG (electroencephalogram), ECG (electrocardiogram), EOG (electrooculogram) are used for intention detection. Among them most commonly used physiological signal for exoskeleton is EMG. Every signal has their limitation but EMG may predict the human movement before occurs. Surface EMG signal has a lot of muscle strength information, easy to get and stable in nature. By using those signals, artificial intelligence algorithms estimate angle, force or torque with continuous smooth control towards trajectory. Siddiqi et al [28] used support vector machine (SVM) technique with puk kernel for thumb angle estimation with an average accuracy of 86.5%. Xin et al [39] has predicted angle estimation of wrist and hand flexion by using TDRNN (Time delay recurrent neural network) methods. Tageldeen et.al. [26] in 2016 used a fuzzy logic technique to estimate joint (elbow) torque from relevant sEMG signals for accurate control of exoskeleton. Zang [32] predicted angle movement intention by BP neural network. Bahareh et al. [37] estimated the joint force of wrist and hand movements with ANN (artificial Neural Network). We receive physical signals from sensors attached to the suit as force or torque sensor, position sensor. A model is created through artificial intelligence that combines physical signals to establish movement intent mapping. Khan et al. [40] utilized extreme learning machine (ELM) to predetermine human motion intentions, here the hidden layer weights were arbitrarily pick out, and the output weights were determined by inverse operation of the hidden layer matrix, so as to reduce the learning time.

Both physiological and physical signals are non-linear types. MLAs has important role for non-linear signal processing. Somehow MLAs has some limitations which needs further improvement. Deep learning process may regain the problems with faster and better recognition.

V. Control Techniques:

An intelligent soft rehabilitation exoskeleton is a combination of bionic structure design and artificial intelligence control algorithm. An exosuit must detect the user's motion and provide information for human-robot interactions that are controlled by an appropriate control algorithm. The human robot interactive control is very complex, where robot compliance takes a big part. Robot compliance may be two types active and passive. Passive one applicable for environmental change and active one for contact force between robot and human. The robotic devices are able to adapt the change as per patient with the combination of intelligent control.

The adaptive controller can adjust some control parameters to accommodate dynamic changes in machine-human interactions. The adaptive controller gives a time-variant model to send feedback to the input to reduce difficulties which occurs due to uncertainty.

utilized reinforcement-based compensation for enhancing the control performance.

Finally, a set of soft wearable robots has been designed for specific functional requirements

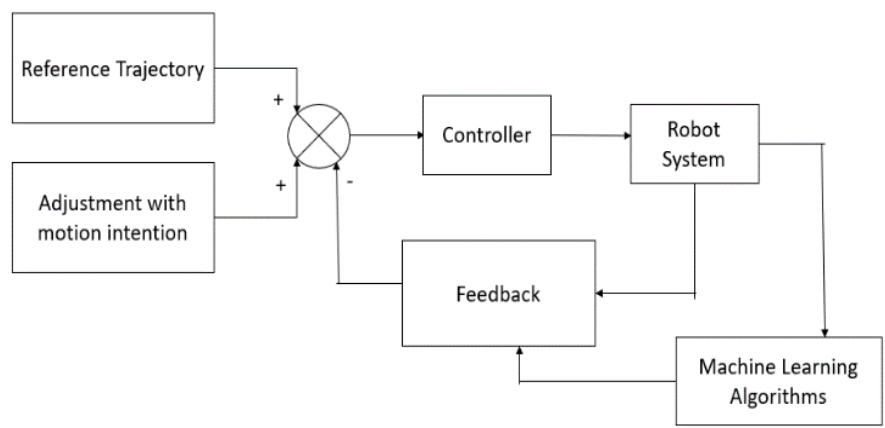

Figure 7: Intelligent controller

The NN based adaptive controller finds optimal solutions after learning from patient's data [41]. Khan et al [40] used a adaptive impedance controller with RBFNN for upper limb dynamic exoskeleton for extracting motion intention. Qingcong et al [41] developed a RBFN-based neural – fuzzy adaptive control strategy for an upper limb rehabilitation exoskeleton. But this type of controller contains some weakness. Hussain et al [44] predicts of the torques of upper limb during eating by use of NARXNN techniques. Yan Chen and others [46] represents a model based on sEMG signal and deep learning method for continuous estimation of upper limb joint angle. Reinforcement learning-based artificial intelligence controllers learn from experience and try to find optimal solutions, but still do not meet the requirements of RL controller researchers. It is not suitable to find the expected goal because it takes more time to avoid unexpected goal. So, RL controller needs more improvement further. Some new technologies needs to input to that type of controller to get ecpected task quickly as optimization, simulation etc. Pane et al [48] to support various human joints. In order to improve human-machine compatibility and assistive performance, related researchers have been trying to create different methods by accumulating rich technological achievements in various processes. The resulting prototypes can contribute exceptionally well to many aspects of people's daily lives.

VI. Assessment:

Rehabilitation robotic devices record the subject's data and make the treatment smooth with assessment process. It gives feedback to therapist and patients to reduce time of the patient's recovery. The quantitive evolution processes determine the state of the patient and evaluate the time requirement for improvement of the patient's movement ability. The clinical assessment tracks the patient's record and the customised therapy makes patient's recovery quickly by physiotherapist. Liao et al. [58] represented a deep learning framework for automatic evaluation, rehabilitation training consists of qualitative, quantitative movement indicator, scoring function, and a deep neural network models for quantifying movement scores. Silvia et al. [59] used wearable sensor data to estimate the scores of FMA (Fugl-Meyer Assessment) with random forest (RF) algorithms. FMA is a simple assessment process conducted by competency-based labor chart-based clinical examination, it is very famous for upper limbs after stroke. The FMA process requires skilled personnel to conduct the tests and sufficient time to conduct the process. Robot-assisted assessment can quantitatively record patient details to restore motor function.

**Discussion:**

The role of the human upper limbs is very important for performing personal activities in daily life. A patient's daily activities can be greatly affected by improper functioning of this organ due to neurological disorders or surgery. The design of the exoskeleton still requires many adaptations, such as moving towards flexible, adaptable, wearable and lightweight structures, as some devices still do not fully meet these criteria in rehabilitation training. Actuated joints mostly have 1 DOF or 2 DOFs. But accurate self-actuated robotic devices require multi DOFs to get good results. The main purpose of using a soft exosuit is portability and ergonomics. Nam et al. [25] paper proposed a lightweight, ergonomic and low-power-demanding soft wearable robot using surface EMG signals to assist the elbow, wrist and fingers. Soft-structures with wire-actuated exoskeletons are also preferred in rehabilitation processes because they are more friendly for daily life due to their lightness and performance, as Thompson et al [38] represented a cable-driven soft actuation system for shoulder and elbow joints for increasing the muscle strength of the operators. Hybrid actuation mechanisms are on the rise these days, giving great satisfaction in certain cases. In [38] Thompson et al proposed a hybrid exosuit that integrates NMES to facilitate whole upper limb muscle strength with motor recovery.

Model-based control is common for operating exosuits. When it comes to higher DOF, more computational power is required and execution takes longer. That's why we want more intelligent soft suits to treat the user's intentions. The user's willingness to exercise is also a major factor in the rehabilitation process. Physiological and physical signals are combined with artificial intelligence algorithms to derive movement intentions, which estimate exercise time for recovery and provide a trajectory to perform the task.

Future research is trending toward creating exoskeletons that actively learn from input experiences and automatically adapt to the environment or user as artificial intelligence provides the basis for developing more reliable systems. A variety of AI techniques are incorporated into the study to find possible solutions among which artificial neural networks are popular. Since ANN has strong non-linear properties, it shows a large effect on continuous motion estimation results. ANN makes robots movement smoother and safer. As chiaradia [55] represents a smooth control exosuits for elbow which detect motion intention by using gravity compensation. Some studies have also become home base for self-rehabilitation after the Covid pandemic [36]. But neural network based adaptive controller has some limitations due to slow execution time for stable and smooth movement of the robot. So that in the future we need deep learning algorithms for control. The quantitative assessment also needs for scoring movement.

Almost all listed robotics exoskeletons are capable of muscle power but are not intelligent enough to function alone. Yet existing intelligent soft exoskeletons are not suitable for clinical use due to some of their drawbacks, with some exceptions. Because existing soft exosuits use very small artificial intelligence controls, the literature shows that there is a need to develop integration methods for intelligent systems and specialized 4hardware that cover user needs in a global, comprehensive way.

**Conclusion:**

The soft exoskeleton is currently a boon for everyday life. It protects our organs from damage. A variety of exosuits have been presented, each of them a new concept and attitude to the device, which has been in the form of prototypes, early clinical studies, and sometimes extensive clinical and commercial research to solve movement disorders and rehabilitation problems. These concepts require more extensive clinical trials to become a complete and efficient plan. For this, related researchers are trying to develop different types of algorithms to achieve different soft upper limb exosuits.

The present scenario of upper limb rehabilitation systems shows that transportable robotic exoskeletons composed of intelligent control and information processing systems can play an important role in bridging the gap between prototype and clinical acceptance. Additionally, exoskeleton capabilities can be greatly enhanced by improving materials and incorporating a better mechanical design, and improving motion sensing techniques.